%% file: MethodPaper.tex
\documentclass[11pt,times,preprint]{elsarticle}
 
\RequirePackage{geometry}
\usepackage{lineno,hyperref}
\usepackage{afterpage}
\usepackage{tcolorbox}

\usepackage [english]{babel}
\usepackage [autostyle, english = american]{csquotes}
\MakeOuterQuote{"}

\modulolinenumbers[5]

\usepackage{subfigure}

\usepackage{xcolor}

\newcommand{\ourparagraph}[1]{\vspace{1mm}\noindent\textbf{#1}}

\input{macros}

\biboptions{authoryear}

\begin{document}

\begin{frontmatter}

\title{BARD: A structured technique for group elicitation of Bayesian networks to support analytic reasoning
}

\author[1]{Ann E. Nicholson\corref{cor1}}
\ead{Ann.Nicholson@monash.edu. Phone: +61 448 019 439}
\author[1]{Kevin~B.~Korb}

\author[1]{Erik~P.~Nyberg}

\author[1]{Michael~Wybrow}

\author[1]{Ingrid~Zukerman}

\author[2]{Steven~Mascaro}

\author[1]{Shreshth~Thakur}

\author[1]{Abraham~Oshni~Alvandi}

\author[1]{Jeff~Riley}

\author[1]{Ross~Pearson}

\author[3]{Shane~Morris}

\author[1]{Matthieu~Herrmann}

\author[1]{A.K.M.~Azad}

\author[4]{Fergus~Bolger}
\author[5]{Ulrike~Hahn}
\author[6]{David~Lagnado}

\cortext[cor1]{Corresponding author}

\address[1]{Monash University}
\address[2]{Bayesian Intelligence}
\address[3]{Automatic Studio}
\address[4]{University of Strathclyde}
\address[5]{University College London}
\address[6]{University of London, Birkbeck}

\begin{abstract}

In many complex, real-world situations, problem solving and
decision making require effective reasoning about causation
and uncertainty. However, human reasoning in these cases is prone to confusion and error. Bayesian networks (BNs) are an artificial intelligence technology that models uncertain situations, supporting probabilistic and causal reasoning and decision making. However, to date, BN methodologies and software require significant upfront training, do not provide much guidance on the model building process, and do not support collaboratively building BNs. 
BARD (Bayesian ARgumentation via Delphi) is both a methodology and an
expert system that utilises (1) BNs as the
underlying structured representations for better argument analysis,  (2) a multi-user web-based software platform and 
Delphi-style social processes to assist with collaboration, and 
(3) short, high-quality e-courses on demand, a highly structured process to guide BN construction, and 
a variety of helpful tools to assist in building and
reasoning with BNs, including an automated explanation tool to assist
effective report writing. The result is an end-to-end online platform, with associated
online training, for groups without prior BN expertise to understand and analyse a problem, build
a model of its underlying probabilistic causal structure, validate
and reason with the causal model, and use it to produce a written
analytic report. Initial experimental results demonstrate that BARD aids
in problem solving, reasoning and collaboration.

\end{abstract}

\begin{keyword}
Probabilistic reasoning; probabilistic graphical models; causal reasoning; collaborative process; Delphi process.
\end{keyword}
\end{frontmatter}


\section{Introduction}
\label{sec:intro}
\input{intro}

\section{Background}
\label{sec:bg}
\input{background}

\section{BARD Approach}
\label{sec:bard_meth}
\input{approach}

\section{Evaluation}
\label{sec:eval}
\input{evaluation}

\section{Conclusions and Future Work}
\label{sec:concl}
\input{conclusion}

\section*{Acknowledgement}

Funding for the BARD project was provided by IARPA through their CREATE project, under contract number 2017-16122000003.


\bibliography{MethodPaper}

\end{document}

%% file: macros.tex
\newcommand{\eg}{e.g., }
\newcommand{\ie}{i.e., }

\newcommand{\PR}{\hbox{Pr}}

\newcommand{\bem}{\bfseries \itshape }

\newcommand{\myparagraph}{\vspace*{-3mm}\paragraph}

\newcommand\code[1]{ {\small{\url{#1}}}}

\definecolor{lgray}{gray}{0.9}
\definecolor{dgreen}{rgb}{0,0.55,0}
\definecolor{purple}{rgb}{0.63,0.13,0.94}

%% file: intro.tex
In many complex, real-world situations, problem solving and
decision making require effective reasoning about causation
and uncertainty. The effectiveness of human reasoning in these
cases is limited: it may handle simple, quasi-linear cases well,
but in complex or non-linear cases it is notoriously prone to
confusion and error~\citep{KahnemanEtAl1982,HahnHarris2014,NewelEtAll2015}. One
way to handle such reasoning more effectively is to employ {\em Bayesian
Networks} ({\em BNs}\/)~\citep{Pearl1988,KorbNicholson2011} which are an
Artificial Intelligence (AI) technology that models uncertain situations, supporting probabilistic and causal reasoning and decision making.

BNs have been deployed for this purpose in diverse
domains such as medicine~\citep{SesenEtAl2013,FloresEtAl2011},
education~\citep{StaceyEtAl2003}, engineering~\citep{BayraktarEtAl2009,
ChoiEtAl2007, MisirliEtAl2014}, surveillance~\citep{MascaroEtAl2014},
the law~\citep{FentonEtAl2013, LagnadoGerstenberg2017}, and the
environment~\citep{CheeEtAl2016}. Furthermore, BNs have been used to analyse common fallacies in informal
logic~\citep{Korb2004}; analyse and assess a variety of arguments
in criminal law, exposing some common errors in evidential
reasoning~\citep{Lagnado2013, FentonEtAl2013}; analyse human
difficulties with reasoning under uncertainty~\citep{HahnOaksford2006,
Hahn2014}, and proposed as a general structured method for argument
analysis~\citep{KorbNyberg2016}.

BNs take advantage of the natural ability of humans to reason and build
causal models about the world~\citep{LagnadoSloman2004, LagnadoSloman2006,SlomanLagnado2015,BramleyEtAl2017}. However, for domain experts to construct their
own BNs, current software usually requires substantial training, 
does not  provide much guidance on the model building process, and does not provide 
any support for collaboratively building BNs. Our system addresses
these deficiencies. BARD ({\em Bayesian Argumentation via Delphi}\/)
combines BNs with a Delphi social process: a systematic method for
combining multiple perspectives in a democratic, reasoned, iterative
manner~\citep{LinstoneTuroff1975}.  The key novel features of BARD,
our structured methodology for group reasoning, are: (1)~customised
Delphi-style BN elicitation; (2)~structured, iterative and incremental
BN building; and (3)~structured, semi-automated narratives.

\myparagraph{\bem Customised Delphi-style BN elicitation}
Analysts in small groups, optionally assisted by a facilitator, are guided
through a structured Delphi-like elicitation protocol to consider and
represent their problem-relevant knowledge in a causal BN augmented by descriptive annotations. BARD provides
tools to assist elicitation of BN structure and parameters, review
and consensus building within the group, and evaluation of the results.
BARD's Delphi-inspired social process helps groups of analysts avoid
common pitfalls. Analysts are required to first
develop an answer on their own, in a private phase, and shows other group
members' contributions after the analyst has published his or her initial
attempt. This approach supports an analyst's modification of their
initial models, while maximising the diversity of answers from which
the group starts its work. The other major features of Delphi utilised
by BARD include anonymity and moderated discussion, both of which help
groups avoid being unduly influenced by the opinionated rather than
the knowledgeable.

\myparagraph{\bem Stepwise, iterative and incremental BN building}
BARD breaks down a given task into six steps that are performed by
the analysts: (1)~pre-modelling exploration of the problem to be solved,
(2--4)~building the components of the BN, (5)~exploring the BN's reasoning
on specific scenarios, and (6)~report writing with BARD's support. However,
progress needn't be linear: BARD encourages analysts to incrementally
and iteratively build their individual BNs and seek regular feedback
through communication with other group members and the facilitator.

\myparagraph{\bem Structured, semi-automated narratives}
BARD guides verbal reporting with an analytical template, designed to elicit relevant points in a logical and thorough way that is consistent with general good reasoning guidelines (\eg~\cite{ICD-203-2015}). BARD also auto-generates from the BN model many key points, in English, organised according to the same template---such as the diagnosticity of evidence and critical uncertainties---which analysts and the facilitator can easily incorporate into their solutions.
\vspace*{2mm}

The development of BARD started as part of the CREATE (Crowdsourcing
Evidence, Argumentation, Thinking and Evaluation) program funded by
IARPA (Intelligence Advanced Research Projects
Activity)\footnote{\url{https://www.iarpa.gov/}}.
The CREATE program sought to develop, and experimentally test, systems
that use crowdsourcing and structured analytic techniques to improve
analytical reasoning, including to help people better understand the evidence
and assumptions that support or conflict with conclusions. CREATE's
secondary aim was to aid users to improve their communication of their
reasoning and conclusions. This included meeting the guidelines for
high-quality analytical reports outlined in the Intelligence Community
Directive 203 (ICD-203)~\citep{ICD-203-2015}. Even though BARD aims to
improve the quality of analysis using BNs, it is not designed for
BN experts. A core purpose of BARD is to make BNs accessible to the
uninitiated, so that those outside the BN community can benefit from
them.

BARD is based on a novel combination of two techniques: BNs and
Delphi (Section~\ref{sec:bg}). The BARD approach presented in
Section~\ref{sec:bard_meth} is the first structured workflow that supports
non-BN experts, with a minimal amount of training, to build and reason
with BNs, with automated explanations. BARD is also the first BN tool
to support a Delphi-style social process that allows participants to
collaboratively construct a group BN solution.  Section~\ref{sec:eval}
summarises results from empirical studies that provide evidence
of BARD's efficacy as a structured technique for improving group
reasoning. Section~\ref{sec:concl} concludes by summarising the
contributions of the paper, and outlines directions for future work.

%% file: background.tex
In this section, we provide background about BNs and Delphi elicitation protocols for group decision making.

\subsection{Bayesian Networks}
\label{sec:cbns}
BNs are the culmination of a century of research on models formally
representing causal relations, beginning with the work of Sewall Wright
in the 1920s and 30s on path models~\citep{Wright1934}. This tradition
has given rise to structural equation models, which underwrite much of
the formal work in economics, psychology and the social and biological
sciences. It also led to work on discovering causal relations from data,
including work by Herbert Simon and Hubert Blalock on ``non-experimental''
causal inference~\cite[\eg][]{Simon1954,Blalock1964}. In the 1980s,
statisticians and AI researchers developed new techniques for modelling
probability distributions, which were codified in Judea Pearl's text {\em
Probabilistic Reasoning in Intelligent Systems}~\citep{Pearl1988}. This
work launched {\em Bayesian Networks} ({\em BNs}\/) as a modelling tool
for reasoning and decision-making under uncertainty, and its theoretical
underpinnings as a field of study.

A BN is a directed, acyclic graph whose nodes represent the random
variables of a problem, and whose directed links (arrows) represent
direct probabilistic dependencies between the nodes they connect~\citep{Pearl1988,
KorbNicholson2011}; in so-called causal BNs, these dependencies are also supposed to be causal. Each node at the tail of an arrow is a {\em
parent} of the node at the head of the arrow (it's {\em child}). The relationship between each child and it's parents is quantified,
for discrete variables (\ie with a finite number of possible states),
by a {\em Conditional Probability Table} ({\em CPT}\/) associated with the
child node. The CPTs of a BN jointly give a compact representation of
the full joint probability distribution of the variables in the BN. Users can set
the values of any combination of variables in a BN, usually on the
basis of observed evidence $e$. This evidence propagates through the
network, producing a posterior probability distribution $\PR(X|e)$ for
each variable $X$ in the network. There are several efficient algorithms for propagating evidence, supporting a powerful combination of predictive,
diagnostic and explanatory reasoning.

\input DrugCheat

Figure~\ref{fig:drugcheat_exampleBN:b} illustrates a simple BN,
where the {\em Drug Cheat} variable represents whether an athlete
has taken performance enhancing drugs, while {\em Sample A Result}
and {\em Sample B Result} represent the results of a test to
detect a performance enhancing drug. Note that the variables are all
discrete 
(Figure~\ref{fig:drugcheat_exampleBN:a}). The arrow from the {\em Event}
variable (Figure~\ref{fig:drugcheat_exampleBN:b}) shows that the probability
of {\em Drug Cheat}={\em True} is influenced by the event, while the
arrow from {\em Taking M879} indicates that the medication M879 may lead to
a positive test result. The combination of influences on the test result
are quantified in the CPT for {\em Sample A Result}, while the different
drug cheating rates for different events are shown in the CPT for {\em
Drug Cheat} (Figure~\ref{fig:drugcheat_exampleBN:c}). The prior probability
for a competitor being a drug cheat, {\em P(Drug Cheat = True)}, is computed to
be 2.33\%, while the sequence of new probabilities computed by the
BN software for Sam the Swimmer is 32.41\% after a positive result for Sample A, jumping to 95.79\% after a positive result for Sample B,
and finally decreasing to 49.24\% in light of new information about taking
M879 medication.

\subsubsection{Probabilistic reasoning errors}
As shown in many studies, human reasoning under
uncertainty is fraught with cognitive biases, which result
in an incorrect update or utilisation of probabilistic
information. Some examples are: {\em overconfidence}---exaggerating
the probability of likely events and the improbability of unlikely
events~\citep{LichtensteinEtAl1982,HealyMoore2007bayesian};
{\em base-rate neglect}---ignoring objective prior
probabilities~\citep{TverskyKahneman1982b,WelshNavarro2012seeing}; and
{\em anchoring}--- depending too much on an initial piece of information
(the anchor)~\citep{KahnemanEtAl1982}.
While many
structured representations may
assist in the avoidance or mitigation of cognitive biases
when analysing problems, causal BNs are particularly suited
to biases involving probability or causality.
By design, BNs only permit logically consistent probabilistic information to be specified in each model, and they provably compute the probabilistic consequences without error or bias. So, if the correct elementary information can be elicited from humans (the arrows, CPTs, and any observational inputs), all the more complex probabilistic calculations will also be correct. Most of BARD's other features (including stepwise construction and Delphi, both described in Section~\ref{sec:bard_meth}) are designed to promote accurate elicitation. 
Accordingly,
the process of modelling reasoning under uncertainty via causal
BNs has been shown to help avoid several common human reasoning
fallacies, such as base-rate neglect~\citep{KorbNyberg2016},
confusion of the inverse~\citep{VillejoubertMandel2002}, the
conjunction fallacy~\citep{JarvstadHahn2011}, the jury observation
fallacy~\citep{FentonNeil2000} and, most recently, the zero sum
fallacy~\citep{PilditchEtAl2019a}.

In addition, people often make reasoning errors in
relation to causality. For example, people often mistake correlation
between events for direct causation, when a hidden common cause may be more
likely~\citep{LagnadoSloman2004, PearlDana2018book,
kushnirEtAl2010inferring, GopnikEtAl2001causal}. Causal BNs
discourage such mistakes, partly because
analysts are forced to think about and model direct causal relations explicitly.
Two examples of causal reasoning phenomena involving indirect
causal connections that are difficult for people to handle, but are
correctly captured by causal BNs, are: {\em explaining away}---when the
confirmation of one cause lowers the probability of an alternative cause
\citep{LiefgreenEtAl2018}; and {\em screening off}---when knowledge of the
state of a common cause renders two dependent effects independent of
each other~\citep{Pearl1988}.

\subsubsection{BN tools}
\label{sec:bns}
Given these benefits, it is not surprising that BNs have been applied
to many application areas, as detailed above, in tandem
with the development of BN software tools that allowed technologists
to build, edit, evaluate and deploy them. 
Widely-used commercial BN software tools include Hugin,\footnote{\url{https://www.hugin.com/}}
GeNie,\footnote{\url{https://www.bayesfusion.com/}}
Netica,\footnote{\url{https://www.norsys.com/index.html}}
AgenaRisk\footnote{\url{https://www.agenarisk.com/}} and
BayesiaLab.\footnote{\url{http://www.bayesia.com/}} In addition, research
software and tools include
Elvira,\footnote{\url{http://leo.ugr.es/elvira/}}
R BN libraries,\footnote{\url{http://www.bnlearn.com/}}
BNT,\footnote{\url{https://github.com/bayesnet/bnt}}
SamIam\footnote{\url{http://reasoning.cs.ucla.edu/samiam/}} and
BayesPy.\footnote{\url{https://pypi.org/project/bayespy/}}

\subsubsection{Elicitation of BNs}
\label{sec:elicitation}
In order to overcome the ``knowledge-engineering
bottleneck''~\citep{KorbNicholson2011}, machine learning methods
for learning BNs from observational datasets were invented, with
several of these algorithms (\eg the PC
algorithm~\citep{SpirtesEta;2000}, CaMML~\citep{ODonnellEtAl2006} and
the R libraries) incorporated into BN tools. However, in the absence of adequate datasets, BNs can be built
through elicitation of domain-specific knowledge from experts. (Expert elicitation and machine learning can also be combined.) Proposed methodologies for constructing
BNs through elicitation are based on concepts such as building BNs iteratively
and incrementally~\citep{LaskeyMahoney1997, LaskeyMahoney2000,
KorbNicholson2011, Boneh2010}, breaking down complex models into
sub-models or fragments, and building BNs with common structures or
elements called ``idioms''~\citep{FentonNeil2014}. However, none of the
commercial BN software packages support the structured elicitation of
BNs, or these knowledge engineering principles. Instead, they assume
that users understand BN technology, and know how to translate their
knowledge of a causal process or argument into a BN. The BARD system
is designed to fill this gap.

The various BN elements that need to be elicited during BN construction
include its variables, arrows, parameters
(conditional probabilities) and their combinations in subnetworks
and networks. There is active research in the acquisition of these
elements,
mostly concentrating on elicitation from domain experts; 
for example, concept mapping is a popular technique for relating
concepts in graphs, which can be used as a preliminary step in
building BNs~\citep{Novak2010}.

Korb and Nicholson have advocated an iterative and incremental approach
to BN construction~\citep[][Part III]{KorbNicholson2011}, suggesting that
models should initially be built up from a small local structure around
a target variable of interest, rather than by attempting to exhaustively
consider every possible factor relevant to the target variable from
the start. In this approach, subsequent iterations pick up a
few additional factors at a time, with validation
(\eg using expert feedback) in each iteration. BARD supports such an
incremental and iterative BN construction approach, where the feedback
is provided by a group of analysts solving the same problem, instead
of an expert.
Overall, BARD's
incremental approach belongs to the ``spiral prototyping'' or ``agile
model building'' family, which adapts ideas long used in software
engineering~\citep{Boehm1988-spiral}.

Probability elicitation has been attacked from a variety of
directions. Since users are sometimes uncomfortable specifying
exact probabilities, even when they are informed that they needn't be
treated as precise, probabilities have often been replaced by language
equivalents~\citep{Elsaesser1987,vanderGaagEtAl1999}---an approach that
was also adopted in the ICD-203 mapping of probabilities to English
(Table~\ref{tab:probabilityDescriptors}). In BARD, we have taken a dual
approach, where probabilities may be viewed both numerically or verbally.

Several protocols have been proposed for eliciting
probability intervals, such as a 3-pt
method~\citep{MalcolmEtAl1959,SollKlayman2004}, a 4-pt
method~\citep{Speirs-BridgeEtAl2010} and the IDEA
protocol~\citep{HemmingEtAl2018a}.
These interval protocols have also been used to combine the responses of multiple experts---and one study used a form of Delphi for exact CPT
elicitation~\citep{EtminaniEtAl2013}. However, to date, these protocols
have not been integrated into any commercial or research BN software
tools. Instead, they are applied outside the
BN software, and their outcomes are incorporated into BNs by the model
builder~\citep{NicholsonEtAl2011,GaagEtAl2012,PollinoEtAl2007a,HemmingEtAl2018b}.

The elicitation of causal structure is a relatively under-explored
area. A generic prototyping approach to address this problem has been
proposed in~\cite[][Part III]{KorbNicholson2011}. Other approaches propose
``idioms''~\citep{FentonNeil2000}, ``templates''~\citep{LaskeyMahoney2000}
and ``network fragments''~\citep{LaskeyMahoney1997} to represent common types of
causal reasoning to be incorporated into problem-specific models when appropriate.
While the BARD training protocol (Section~\ref{sec:BARD_training})
introduces these concepts, and describes how BARD can be used to
construct models that capture them, there is not yet any explicit functionality to facilitate this
in the tool. Serwylo~\citep{Serwylo2015}
pioneered using online crowdsourcing and automated aggregation for
BN structure elicitation, albeit not in the Delphi style, while~\cite{NicholsonEtAl2016} explored Delphi elicitation
and automated amalgamation of structure and parameters. However, BARD is
the first tool to apply Delphi to the full BN model building process.

During elicitation, it is an important but often under-appreciated task
to document how the model was constructed, \eg the sources of modelling
elements and their reliability. Although as yet there is no accepted
standard for this kind of meta-documentation, BARD offers an initial
approach for structured recording of this meta-information (see Section~\ref{sec:bard_meth}).

Validating computer models means testing their accuracy in
representing a real-world system. In general, computer models are
validated using expert feedback, data or a combination of the
two~\citep{FloresEtAl2011,KorbEtAl2013}. The BN software packages listed
above support data-driven validation in
several ways, but do not support structured expert validation, such as the validation framework proposed by Pitchforth and Mengersen \citep{PITCHFORTH2013162}, beyond
ad hoc exploration and validation of ``what if'' scenarios, as does
BARD. In the BARD environment, group deliberation offers a further form
of validation, \ie via the social process described in
Sections~\ref{sec:workflow} and~\ref{sec:socialprocess}.

\subsubsection{Explaining Bayesian Networks}
\label{sec:explanation}
The automatic generation of explanations
from BNs has been investigated since the early
1980s~\citep{Boerlage1983,SemberZukerman1989,Suermondt1992},
followed by work conducted by authors Korb and
Zukerman~\citep{korbEtAl1997cogsci,ZukermanEtAlAAAI98,ZukermanEtAlIJCAI99,JitnahEtAl2000-towards},
and more recently by other researchers~\citep{Vreeswijk2005,Keppens2011}.
However, automated explanation from BNs has thus far been selective
and special purpose, with language tailored to specific variables and
subnetworks. The algorithms generated to date have achieved achieving
limited success in explaining complex non-monotonic relations.  BARD's
explanation-generation component circumvents some of these problems,
producing explanations that harness the explicit causal nature of
links, as well as common idioms for expressing probabilistic and causal
relationships (Section~\ref{sec:6steps}).

\subsection{Delphi protocols for group decision making}
\label{sec:delphi}
There is considerable evidence that decision making by
groups (either by reaching consensus or amalgamation)
can produce better outcomes than decision making by
individuals~\citep{SalernoEtAl2017Group, KuglerEtAl2012groups,
CharnessSutter2012groups, StrausEtAl2011group}. However, there are
also well-known problems while working with groups, \eg anchoring on the earliest responses,
groupthink, and the excessive influence of higher-ranking members~\citep{KahnemanEtAl1982,
StettingerEtAl2015counteracting, MumfordEtAl2006errors,
Packer2009avoiding}.
Several methods have been developed over the years that attempt to
harness the positives of groups, while preempting or mitigating
the negatives;
one of the most well-established is the Delphi
technique~\citep{LinstoneTuroff1975,RoweEtAl1991}.

Delphi is an example of a
nominal group technique, where the group members never actually meet face-to-face,
but interact remotely. 
Thus, participants need not be present at the same location or make their contributions at the same time---practical benefits when experts are dispersed, perhaps internationally, with limited time and conflicting or busy diaries.
Furthermore, the group members don't even know who
their fellow group members are---a deliberate ploy designed to ameliorate
cues related to supposed seniority, experience or expertise, which may
be unhelpful (as expertise and advancement can often be related
to personality or background characteristics, rather than skill or
knowledge).
Thus, members can focus on the information provided
by others, and the undue influence that powerful or dogmatic individuals can have on group judgments is reduced.

These anonymous participants are first asked to provide their own judgement on the issue at hand, before finding out about the responses of others. This increases the independence and diversity of initial responses, reducing social loafing and the premature conformity seen in anchoring and groupthink. The responses are collated by a facilitator, then fed back to the participants for a second round. The participants consider the information (which may simply be the mean or median of the group response when quantitative values are in question, but may also include rationales/justifications for answers), then provide another response, which could be the same as before, or could be an amended one. This encourages participants to rationally reconsider their response in the light of any new information provided by others. Several rounds may take place, continuing until some stability is achieved (although most changes take place in the second round, and few studies go beyond two or three rounds). This process tends to increase the level of consensus in the group, but the more fundamental aim is to increase the overall quality of the responses. After the final round, the facilitator usually aggregates the responses of the individual members (or collates them, if responses are qualitative in nature), and the resultant answer is taken as the group response. Answers are usually weighted equally, which ensures that the final response reflects fairly the views of all group members. In addition to their benefits for administration and collation, using a facilitator tends to encourage constructive contributions from members and avoid any unproductive, heated arguments.

In summary, the defining characteristics of a Delphi process~\citep{RoweEtAl1991} are: {\em anonymity}, {\em iteration} followed by {\em feedback}, and {\em aggregation} (or collation) of group responses, which is often completed by a facilitator. A review in~\cite{RoweEtAl1999} found that, at least for short-term forecasting problems and tasks involving
judgements of quantities, Delphi has generally shown improved performance compared to freely inter-
acting groups or a statistically aggregated response based on the first-round responses of individual
participants. However, Delphi has not previously been tested on complex reasoning problems.

One Delphi variant is  a ``roundless'' version, called {\em Real-time
Delphi}~\citep{GordonPease2006},  where the iterative process (providing
individual responses, viewing information from other participants,
and amending responses) is not controlled by a facilitator, and for each participant the
transition between steps occurs immediately, \ie in real-time. This
setup is more flexible than regular Delphi in the timing
of participants' contributions, and has the potential to speed up the
Delphi process.
However, since participants can see any other available responses directly and asynchronously,
rather than after amalgamation or collation by a facilitator, some of
the biases associated with direct interaction may re-emerge. The social
process in the BARD methodology is a version of Real-time Delphi. We discuss
our reasons for trading off speed and ease of use against bias in Section~\ref{sec:workflow} below.

%% file: DrugCheat.tex
\begin{figure*}[p]
{\small \sf
\begin{center}
\textsf{\textbf{The Drug Cheat Problem (BARD Training Problem)}}
\end{center}
After competing, a proportion of competitors at the Olympics are randomly chosen for testing for the presence of steroids. Here we'll consider only competitors from three sports: athletic runners, swimmers and weightlifters. Drug tests conducted in the past indicated that 4\% of Weightlifter take performance enhancing drugs, while Runners are half as likely as Weightlifters to take performance enhancement drugs and Swimmers are half likely as Runners. The error rates are 2\% false positive and 5\% false negative. When an athlete is chosen for drug testing, two samples are taken, the A and the B sample, with the B sample only analyzed if the A sample comes back positive. The threshold for being found guilty, which will result in automatic disqualification and a 2 year ban, is 98\%.\vspace*{2mm}

Consider the scenario of a swimmer, Sam, who is randomly chosen for
testing. Sam returns a positive result for first the Sample A Test, and
then the Sample B Test. Sam claims that the positive test result was not
caused by a performance enhancing drugs, but by taking a medication, M879,
prescribed by her doctor that is on the approved list. M879 has recently
been found to trigger a positive result in the test for performance
enhancing drugs.  Sam's doctor
confirms that Sam did take this medication for a condition that is very
rare. Based on the given information and evidence should Sam be found
guilty and hence disqualified and banned for 2 years.\vspace*{1mm}

\centering
\hspace*{-4mm}\begin{tabular}{p{0.50\textwidth} p{0.4\textwidth}}
\subfigure[Variables and their states]{\label{fig:drugcheat_exampleBN:a}
\footnotesize
\hspace*{-3mm}\begin{tabular}{l l}\hline
    Variable & States \\ \hline
    Event & \{weightlifing, running\} \\
    Drug Cheat & \{true, false\} \\
    Testing performed\hspace*{-2mm} & \{yes, no\} \\
    Sample A result & \{positive, negative, noResult\}\hspace*{-2mm} \\
    Sample B result & \{positive, negative, noResult\}\hspace*{-2mm} \\
    Taking M879 & \{yes, no\} \\ \hline
  \end{tabular}
}\hspace*{-4mm} &
\subfigure[Structure]{\label{fig:drugcheat_exampleBN:b}
  \includegraphics[width=0.45\textwidth,height=35mm]{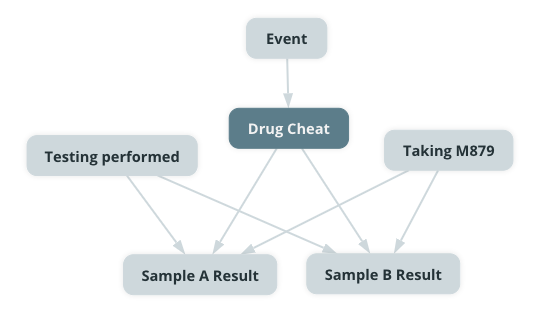}
}
\end{tabular}\vspace*{-3mm}\\
\hspace*{-3mm}\begin{tabular}{p{0.5\textwidth} c p{0.5\textwidth}}
\subfigure[CPTs for {\em Drug Cheat} and {\em Sample A Result} variables]{\label{fig:drugcheat_exampleBN:c}
\begin{tabular}{c}
  \includegraphics[width=0.5\textwidth]{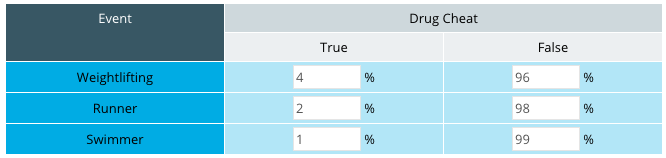}\\
  \includegraphics[width=0.5\textwidth]{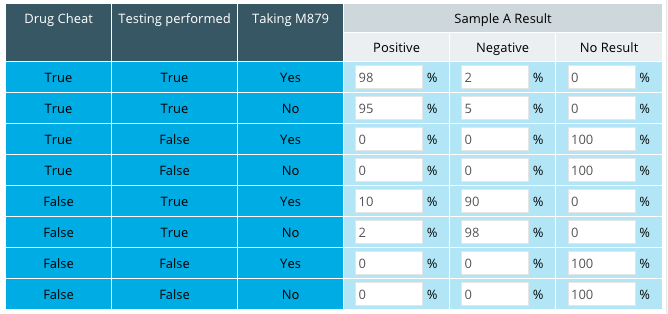}
 \end{tabular}
} & &
\subfigure[Updated probabilities for base case and after each new piece of evidence]{\label{fig:drugcheat_exampleBN:d}
\footnotesize
\begin{tabular}{p{18mm} l}
\raisebox{3mm}{Base scenario} & \includegraphics[width=0.18\textwidth]{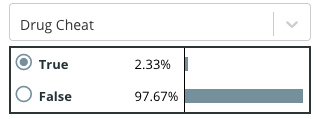}\\
\raisebox{3mm}{\begin{minipage}{18mm}{Sample
A\\Result=Positive}\end{minipage}} & \includegraphics[width=0.18\textwidth]{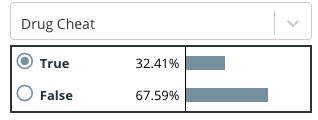}\\
\raisebox{3mm}{\begin{minipage}{18mm}{Sample
B\\Result=Positive}\end{minipage}} & \includegraphics[width=0.18\textwidth]{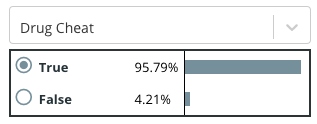}\\
\raisebox{3mm}{\begin{minipage}{18mm}{Taking\\M879=True}\end{minipage}} & \includegraphics[width=0.18\textwidth]{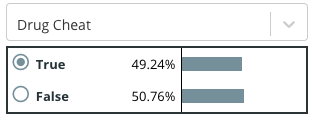}\vspace*{1mm}\\
\end{tabular}
\normalsize
}
\end{tabular}
}
\vspace*{-2mm}
\caption{The Drug Cheat Problem: The BN structure, the CPTs together with base scenario (no evidence) and the updated posterior probabilities for the scenario involving Sam the Swimmer after each new piece of evidence.}
\label{fig:drugcheat_exampleBN}
\end{figure*}

%% file: approach.tex
In the Delphi-style BARD structured group technique, individual group
members (called analysts in BARD) submit their contributions to a problem
anonymously. A moderator (called the facilitator in BARD) guides and
supports the analysts through the process, and ensures an overall group
solution is produced. A BARD group consists of a single facilitator and
any number of analysts.

\begin{figure*}[t]
    \centering
    \includegraphics[width=0.9\textwidth]{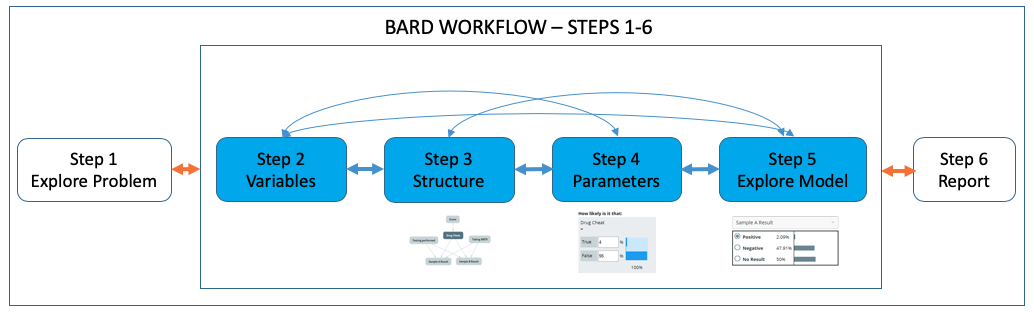}
    \caption{BARD consists of six steps. Analysts and the facilitator can move flexibly backwards and forwards between steps to update their work as necessary.}
    \label{fig:6steps}
\end{figure*}

\subsection{BARD Workflow}
\label{sec:workflow}
The BARD workflow consists of six steps, broken down into three phases,
as depicted in Figure \ref{fig:6steps}. The first, the pre-modelling
preparatory Step 1, focuses on helping the group understand the problem
to be solved and the questions to be answered, along with the main
hypotheses and pieces of evidence. The second phase consists of Steps
2--5, where the focus in on building a causal BN that models the problem
situation and using the causal BN's reasoning to assist in answering
the questions. These steps reflect the natural sequence of tasks in
BN construction: selecting the variables (Step 2), determining the
network structure (Step 3), parameterising the model by eliciting the
CPTs (Step 4), and then exploring the completed model's reasoning on
specific scenarios (Step 5). Step 6 involves production of a structured
written report.  While there is a natural sequence to the workflow, it is
not a one-way street: users can always go back to revise their previous
work.  This supports building the BN iteratively and incrementally,
per best-practice BN construction (Section~\ref{sec:elicitation}).

\begin{figure*}[t]
    \centering
    \includegraphics[width=0.9\textwidth]{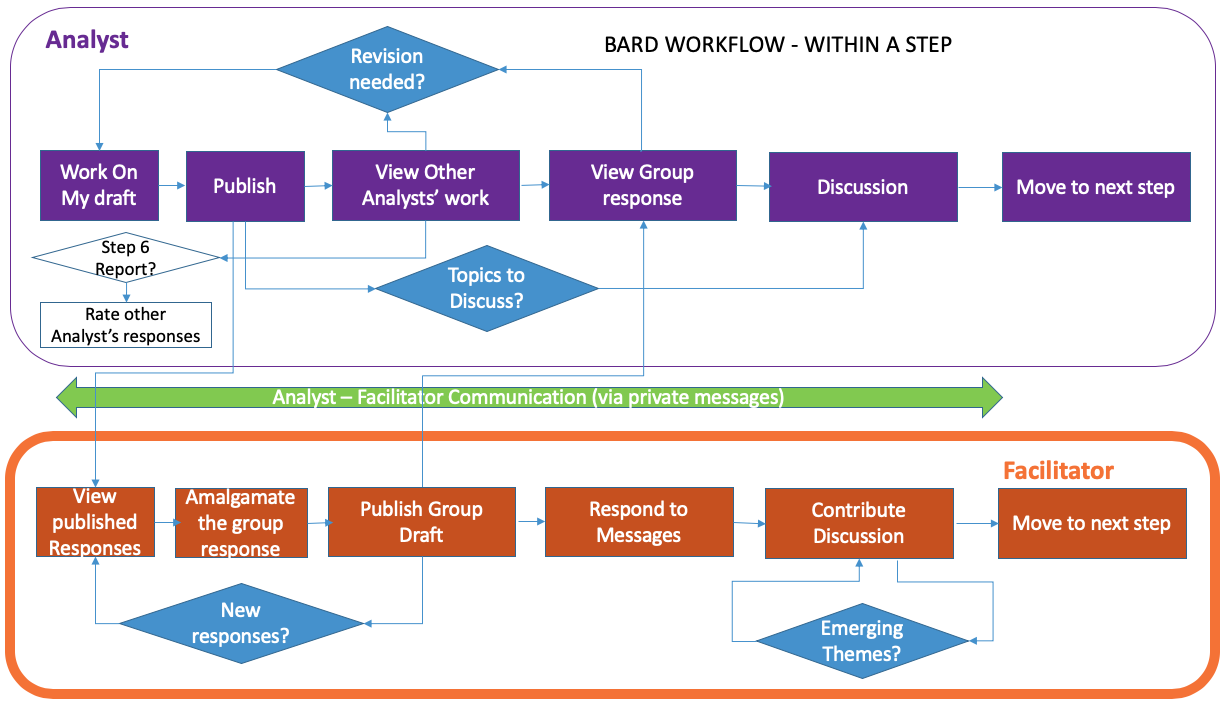}
    \caption{High-level representation of the BARD workflow within a step for analysts (above) and facilitator (below) working in a BARD group.}
    \label{fig:bard_delphi}
\end{figure*}

Analysts contribute their individual domain knowledge and problem solving abilities across these six steps of BARD, while the facilitator constructs a Group version for each step, based on the analysts' work. This is done via a structured workflow {\em within} each BARD step. At each step, analysts are required to first work on their own, and then share that initial attempt with the group. After this, they can view other analysts' work and the current group solution (Figure~\ref{fig:bard_delphi}), discuss solutions via the step-specific discussion forum, and move on to the next step whenever they choose (the ``real-time'' element). Analysts can also move back to an earlier step to revise their work at any time, and then move forward again to any step they previously reached. 

The facilitator's workflow is more flexible than the analysts', as they can move to any BARD step and can view all analysts' shared work at any time. It is the facilitator's role to synthesise the group's work at each step as necessary to develop a coherent solution that reflects the group's thoughts.
They encourage analysts to resolve any points of disagreement themselves; however, they are empowered to make the final decisions, which may involve either
collating different analysts' work or adopting a single analyst's work. The facilitator can present the current group solution (for any or all steps) back to the group at any time so that analysts who have shared their work for that step can view it, discuss and provide feedback, and revise their work if they wish.

The social process within BARD's structured workflow is a flexible variant of Real-time Delphi, arising from our observations during the interactive design and prototyping approach we used during BARD's development. The BARD refinements of Delphi elements reflect the needs we identified to keep participants in the CREATE program engaged and actively contributing in a modern, online environment, including:
(1) the contribution required of group members by BARD---specifically, building and using a BN for relatively complex problem solving---is much more demanding than the tasks typically considered within a traditional Delphi processes;
(2) participants have expectations of significant autonomy 
(rather than having to wait for others at every step), 
while still being able to access peer and facilitator support when needed; (3) participants often have a preference for direct engagement with others (rather than having all communications and information come through a moderator/facilitator);
(4) the BARD process becomes far too drawn-out with multiple Delphi rounds for each of the six steps, especially when the group is working in a distributed, asynchronous manner and may be spread over different time zones;
(5) when one analyst wishes to revise their work from a previous step it is too onerous to require everyone else to return to that step;
(6) groups must be allowed to continue operating in the temporary or permanent absence of a facilitator; and
(7) not all participants will be equally engaged over the whole problem-solving time period, so to maximise their opportunity to contribute when they are engaged, it is beneficial to allow them to complete and comment on earlier steps others have already passed through and/or later steps others have not yet reached (rather than only accepting contributions to a single, current Delphi round).

A further issue we found is that participants vary in their domain expertise or problem solving ability. Hence, while the workflow must support some form of aggregation or collation, BARD does not enforce traditional Delphi equal weighting of all group member contributions for quantitative judgements. Further, for more qualitative judgements, traditional Delphi allows for further rounds until there is no significant change in opinion. This kind of stopping rule may also be used in BARD, but for our experimentation completion of the workflow was deadline driven, and this is likely to be more applicable to future real-world applications.      

Although analysts' real names are concealed, they are assigned pseudonyms that they keep throughout the problem. This helps to identify the work and comments of each analyst in each step, and also relate it to their work and comments in other steps, which makes discussion and comparison far easier for participants.

Although we have deviated from traditional Delphi processes,\footnote{During BARD's development, we prototyped and evaluated other versions of Delphi, two of which are still supported and configurable via the administration panel at problem setup time. These are (1)~a classic Delphi process, with multiple Delphi rounds for each of the six steps of BARD, analysts all moving to the next step at the same time when the facilitator gives them access, no discussion forums, and where analysts only see the amalgamated group version provided by the facilitator rather than each other's work directly; (2)~a variant Delphi process (sitting between the classic Delphi and the default real-time version) where, in addition to the group solution, analysts can view and discuss other team member's published work, but with the facilitator controlling access to the next step.} 
BARD retains some of the key features that have made Delphi a successful method---anonymity, individual judgement before sharing, iteration and feedback, use of a facilitator---with some aspects supported and encouraged rather than enforced. Hence, we describe the BARD workflow as Delphi-style, rather than the Delphi method per se. 

Finally, throughout the six steps, BARD encourages its users to enter a rationale to explain their analysis, making it easier for other group members to understand the solution. This detailed documentation 
improves the exchange of ideas and provides a basis for discussion on points of disagreement, 
and hopefully leads to a better understanding of the problem and the resulting solution.

\subsection{The Six Steps of BARD}
\label{sec:6steps}

Here each of the six steps are described in more detail. 

\ourparagraph{Step 1: Explore Problem} allows analysts to read and examine the problem, review any questions that have been posted, and encourages them to extract key features by identifying (1) the hypotheses suggested by the problem/questions, and (2) the items of evidence most pertinent to those hypotheses. Analysts are also encouraged to provide rationales for the inclusion of each hypothesis and evidence item. The motivation behind this step, as a precursor to the BN modelling, is to have the group gain and record a shared understanding of the problem they must solve, and reach some level of agreement on the key elements that must be included or addressed in BN construction to answer the questions posed.

\ourparagraph{Step 2: Variables} is where the variables of the BN are specified, with BARD suggesting that analysts consider converting the hypotheses and evidence items from Step 1 into variables. BARD prompts for two kinds of variables: {\em Target} variables and {\em Other} variables. Target variables are often the hypotheses or further variables closely associated with the questions to be answered. In BN modelling methodologies, these are also described as ``query'' or ``output'' variables, and these modelling methodologies suggest identifying them first and then focusing attention on variables that are either causes or effects of those target variables---the ``other'' variables in BARD. Target variables are distinguished with a different colour in the Structure visualisation, but all variables (Target or Other) may be the output variables of a scenario in Step 5 Explore Network (see below).  

BARD variables must be specified together with their discrete states,
which BARD currently supports in four categories: Boolean (with just
the two states {\em True} and {\em False}) for propositional variables, \eg {\em
Testing performed} in the Drug Cheat example; Binary (any other two-state
variables), \eg {\em Taking M989}; Ordered (any multi-state variables with
the states in ranked orders), \eg \{{\em High, Medium, Low}\}; and Unordered
(any other discrete variable), \eg the {\em Event} variable has the states
\{{\em Weightlifting, Running, Swimming}\}).

Descriptions of the variables and the variable states are solicited, but
not required, as are ``rationales'' for the choice of variables. These
meta-data items are intended not only to document the intent and meaning
being these modelling elements, but also to stimulate active discussion
when other analysts see them and disagree.

\ourparagraph{Step 3: Structure} is where the relationships between the variables are specified. In this step, BARD displays each variable as a draggable node (with its name)
on a canvas, and prompts analysts to specify the causal 
structure by drawing \textit{arrows} between pairs of variables, i.e., graphically specifying relationships to produce a node-link (or ``network'') diagram for the BN.  Target variables are differentiated
from other variables by colour (Figure~\ref{fig:drugcheat_exampleBN}).
Arrows can be readily deleted or redirected to a new variable. Analysts can associate text labels with arrows, as well as create general labels anywhere on the canvas to act as titles or general purpose on-canvas documentation (as standard in most BN software GUIs).  

\begin{figure}[tb]
    \centering
    \includegraphics[width=0.45\textwidth]{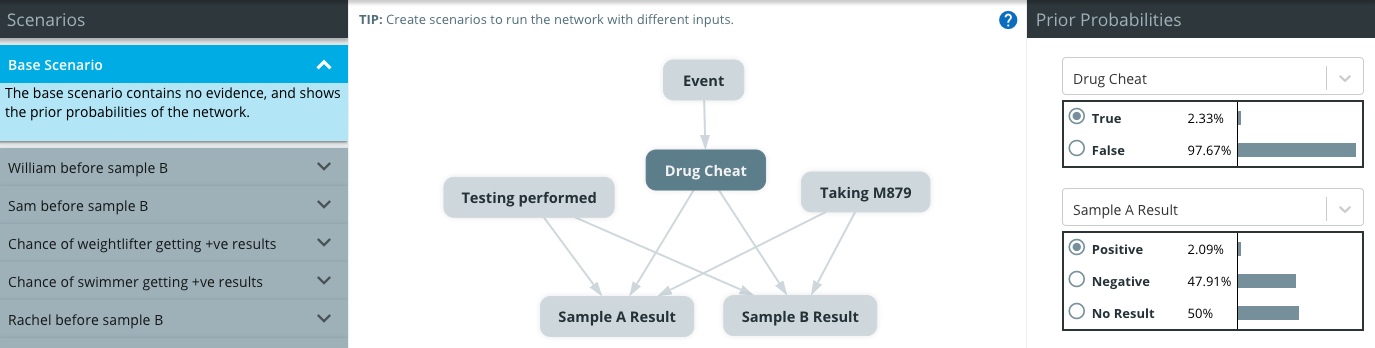}\\
    \includegraphics[width=0.45\textwidth]{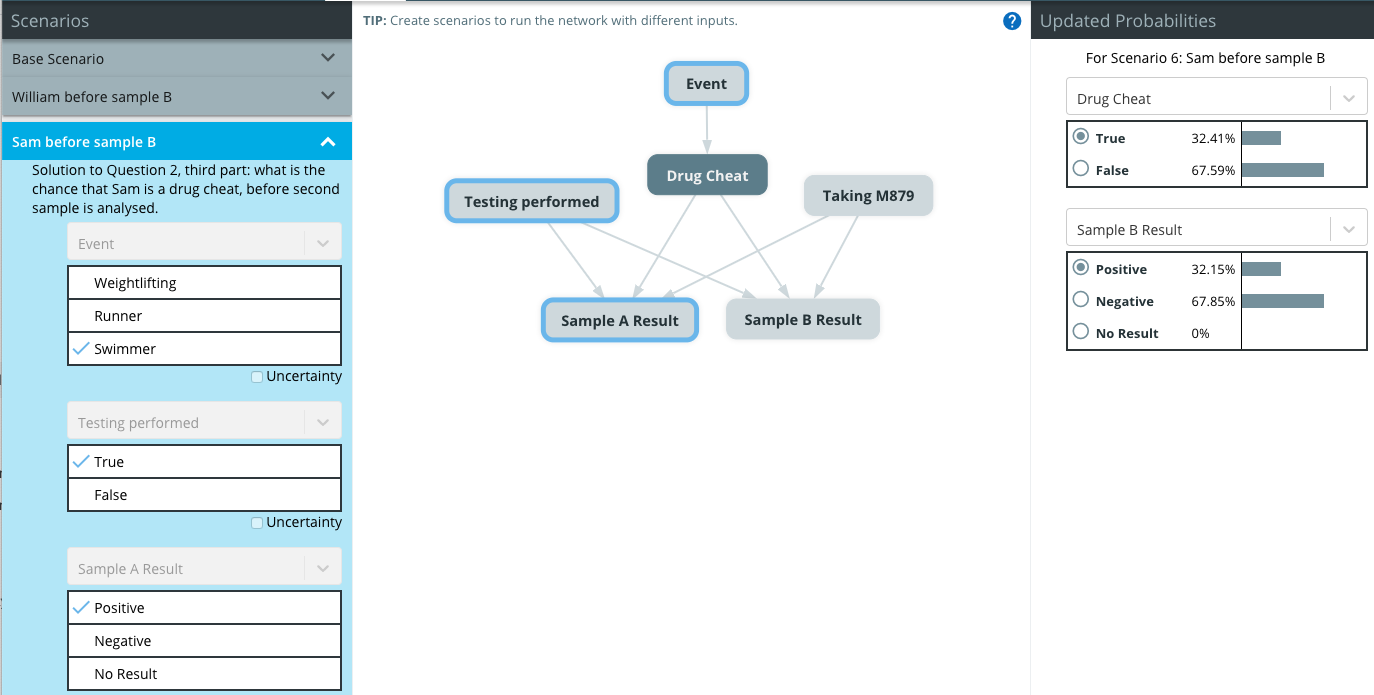}\\
    \includegraphics[width=0.45\textwidth]{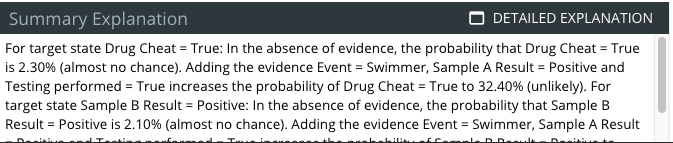}
 
    \caption{Step 5: Explore Network allows evidence to be added into scenarios in the left panel and the resulting updated probabilities in the right panel. The model in the centre panel provides an overview of the network and highlights the evidence variables in blue. The screenshot above shows the base scenario, along with a scenario after each new piece of evidence is added. Below is summary explanation for the third scenario (with all three pieces of evidence).}
    \label{fig:step5_eg}
\end{figure}

During this step and later steps that display a network view of the BN
(\eg Step 5 as shown in Figure~\ref{fig:step5_eg}), BARD adjusts the
layout of the network to enforce a natural causal ``flow'' (left-to-right
and up-to-down) and to prevent graphical elements from overlapping.
This is achieved using a technique called constraint-based layout
(CoLa)~\citep{DwyerEtAl20090-cola}.\footnote{Currently, the configuration
of these layout constraints is done in the software. In future, we plan
to make this configurable by the BARD administrator or by users.} As
the analyst specifies arrows, CoLa automatically shifts variables around
the canvas to maintain distance between them, prevent variable overlaps
and minimise overlap between arrows and labels.  One advantage of this
automation is reduced effort by the user, but also, it's easier for users
to recognise similarities and differences in other models when they are
laid out in the same way.  CoLa does allow analysts to reorganise the
network layout manually, by clicking and dragging variables around the
canvas, while still enforcing some layout constraints.

At this point, the analyst may wish to add additional variables or
modify the states or names of existing variables. They can do by
returning to Step 2, per the flexible BARD workflow across steps
(Figure~\ref{fig:6steps}).

\begin{figure*}[t]
    \centering
    \begin{tabular}{|c|c|}
 \hline
 \includegraphics[width=0.46\textwidth]{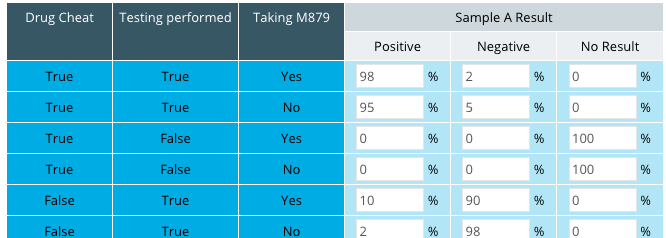} & \includegraphics[width=0.46\textwidth]{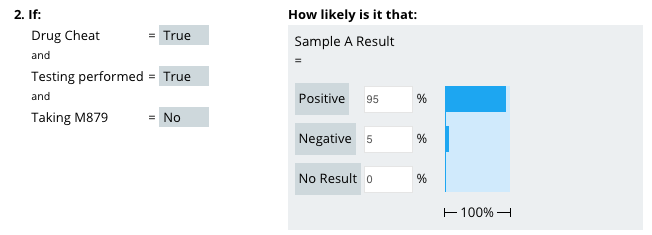}\\
\includegraphics[width=0.46\textwidth]{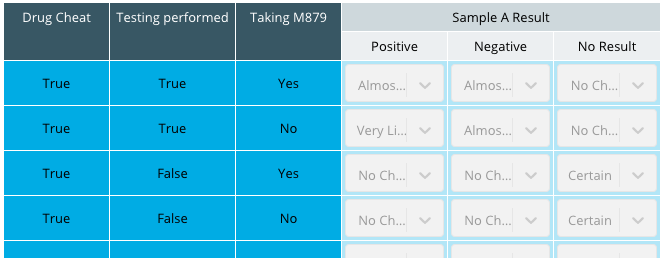} & 
\includegraphics[width=0.46\textwidth]{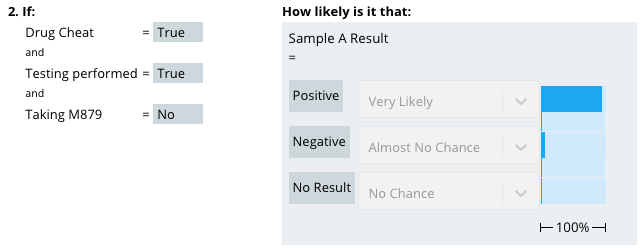}\\
\hline
\end{tabular}
    \caption{Examples of the four input modes available in Step 4 Parameters: percentages above, qualitative descriptors below; table left and question-based right.}
    \label{fig:step4_input_modes}
\end{figure*}

\ourparagraph{Step 4: Parameters} allows users to specify the
conditional probabilities for each child variable given each joint
state of its parents, i.e., the child's CPT. BARD provides two modes for
specifying the conditional probabilities: (1)~as answers to questions,
one question for each combination of the parent node states; or
(2)~via a table.\footnote{We followed the Netica table layout rather than Hugin/Other software, which reverses the rows and columns, because it makes scrolling vertical rather
than horizontal when the table size increases.} In either case, BARD
provides two ways of entering probabilities: (1)~as percentages;
or (2)~as English language verbal descriptors, each of which has an
associated probability range, as given in ICD-203~\citep{ICD-203-2015}
(Table~\ref{tab:probabilityDescriptors}). In combination, this yields four possible
input modes, as depicted in Figure~\ref{fig:step4_input_modes}. This
approach caters for both users who prefer to model with precise parameters
and users who prefer to avoid ``false precision'' by using qualitative
verbal descriptors (as described in Section~\ref{sec:elicitation}).

\begin{table}[t]
\footnotesize
\centering
\begin{tabular}{|c|c|}
\hline
No Chance & 0\% \\ \hline
Almost No Chance & $0< p \leq 5$\% \\ \hline
Almost No Chance & $5\% < p \leq 20$\% \\ \hline
Unlikely & $20\% < p \leq 45$\% \\ \hline
Roughly Even Chance & $45\% < p \leq 55$\% \\ \hline
Likely & $55\% < p \leq 80$\% \\ \hline
Very Likely & $80\% < p \leq 95$\% \\ \hline
Almost Certain & $95\% < p < 100$\% \\ \hline
Certain & 100\% \\ \hline
\end{tabular}
\normalsize
\caption{Mapping verbal probability descriptors to probability ranges, taken from ICD-203~\citep{ICD-203-2015}}
\label{tab:probabilityDescriptors}
\end{table}

Where individual probabilities haven't been specified, the BARD default is a uniform distribution of any unused probability mass. This allows quick specification of CPTs for which only a proper subset of the parameters are known.

\ourparagraph{Step 5: Explore Network} is where the group members can use the BN for reasoning,
thus exploring the consequences of steps 2-4. Evidence 
is added by setting one or more variables to particular states, and the BN reasoning engine computes new probability distributions for the remaining variables. In BARD, each set of evidence is called a ``scenario'' (following AgenaRisk terminology), and may involve setting values for any number of variables. A scenario may describe a specific situation given in the problem description or just a hypothetical ``what-if'' scenario that the analyst wants to explore. In BARD, scenarios are named, can be given associated descriptions, and may be shared and discussed. 
When viewing other analysts' BNs, a BARD user can't edit them, 
but they can explore their consequences by adding new scenarios that are only visible to them. 
Analysts can always change/extend either their own BN or
the associated scenarios, and facilitators can do this for the group's model.
Step 5 always includes a default ``base'' scenario, which shows the probability distributions for all specified output variables when no evidence has yet been added. 

Step 5 is where the group decides whether to continue the spiral prototyping
of the BN or,if they are satisfied with it, move on to Step 6 (Report).
Scenarios allow an explicit and visual way of investigating
the appropriateness of the BN---both its structure and parameters---and
determining whether it is giving a reasonable representation of known
or hypothetical scenarios. Scenarios can also provide a direct means
of answering questions about the confirmatory value of evidence or the
final probability of some event given any combination of evidence.  More formally,
by allowing scenarios to be set up, stored and examined, BARD supports
the following validation activities~\citep{KorbEtAl2013}: {\em
face validity}---checking whether a model captures the known features
of a situation; {\em content validity}---checking whether the model's
confirmatory or causal relationships capture known relations; {\em
case analysis}---seeing whether known cases are modelled correctly;
{\em sensitivity analysis}---determining whether variations in target
variables are proportionate to variations in evidence, including examining
the confirmatory power of different evidence sets. This is obviously
useful for testing whether parameters (conditional probabilities) are
sensible, but can also reveal missing causal connections, for example. In
the future, we anticipate providing more targeted sensitivity analysis
tools, such as reporting Bayes Factors (for confirmation) or causal power
(\eg via the measure in \cite{KorbEtAl2011}).

The BARD Step 5 workspace is divided into three panels
(Figure~\ref{fig:step5_eg}): the left-most panel contains the
scenarios, with the base scenario listed first, and a single active
scenario (selected and expanded) at a time; the middle panel shows the
BN structure; the right panel shows the output variables 
(a subset of all the variables in the BN, as selected by the user) together
with the computed probability distribution over the states; and a summary
explanation is shown below. When a scenario is active, the distribution of
non-evidence nodes may be easily examined. If the user wants to compare
the outputs of two (or more) scenarios, they can either click back and
forth (like AgenaRisk) or open (in another browser window) 
an additional instance of BARD for the
same problem in the same step, to view the
two scenarios side-by-side.

BARD Step 5 includes a general-purpose automated BN explanation tool,
implementing a mix of traditional and novel natural language generation
techniques and taking advantage of the explicitly causal nature of
the links and common idioms for expressing probabilistic and causal
relationships.  The BARD Automated Explanation Tool (AET) can be used
by analysts when exploring a complete causal BN, either to critique the
model, or to contribute to writing up a report based on the reasoning
provided by the BN.

The BARD AET generates both a summary explanation, which is re-generated
and displayed each time probabilities are updated during the Explore
Network step, as well as a detailed explanation that can be accessed by
clicking through to a separate dialog box. For both, target variables
and states (specified directly in the Explore Network step) are used to
focus the explanation. Probabilities are stated both numerically and with
verbal descriptors, following the ICD-203 recommendations
(Table~\ref{tab:probabilityDescriptors}).
The BARD summary explanation provides information
about what the model probabilities would be if no evidence were entered,
specifying only the target variables and states. If the scenario specifies
evidence, additional statements are provided about what the evidence is
for the given scenario, and how the model's probabilities change when we
take this evidence into account. An example of the summary explanation
is given in Figure~\ref{fig:step5_eg}. The BARD detailed explanation
includes multiple elements:
the causal structure of the model, the probabilities of the target variables without any evidence and how the target probabilities are related to each other, the general reliability and bias of the evidence sources, why the evidence sources are structurally relevant, the probabilistic impact of the evidence items on each hypothesis (which can be presented in several ways and include additional notes to highlight interactions between them), and the final probabilities of the hypotheses given all the evidence~\citep{ZukermanEtAl2019}.

The AET has been tested on all ten BARD training problems
(Section~\ref{sec:BARD_training}), as well as the four problems in the BARD empirical
experiments described in Section~\ref{sec:eval}, and in three
additional problems developed for CREATE. Overall, it has been shown
to produce satisfactory English language descriptions. However, we have
not yet empirically tested to what extent the provision of automated
explanations, or which elements of them, improve analytic solutions.

\ourparagraph{Step 6: Report} provides an environment in which the group
can develop a joint written answer to questions raised in the problem
statement. In our preliminary testing, we identified that providing
a template to assist users in writing effective reports improved
their performance. The template encourages analysts to methodically organise and explain their
analysis in detail, and prompts them to include various key elements of good reasoning, such as probability estimates
for key hypotheses in any BN developed. The Automated Explanation Tool
available in Step 5 also provides its detailed output in sections of text aligned with the
template sections, which can be used directly or paraphrased to complete
corresponding sections of the template.

Step 6 also allows analysts to rate final reports, which will either
decide or inform which report is selected as the solution for that
problem. This element was added when preliminary usage indicated that
discussion forums did not always generate a clear consensus or guidance
for the facilitator on the best BN or final report.
Furthermore, in some experiments a few facilitators were no longer active at this stage in the process, so analyst ratings allowed the group report to be selected automatically in these cases.
Rating is done on a
scale from 1 to 10 using a slider. After they have submitted their ratings,
analysts can see the average score for each option and how many ratings have been submitted
so far, but cannot see other analysts' ratings.

BARD has a notion of submission for the final group solution. 
In Step 6, the facilitator can click a `submit' button, which generates a PDF of the group report, and sends it to a nominated electronic location. This
is useful for experiments and for usage when there is a hard deadline,
but not required. All group members have access to the final published
group solution and can download both the written report in PDF format and the BN.

\subsection{The BARD Social Process: roles and communications}
\label{sec:socialprocess}

A BARD group consists of a single facilitator and any number of analysts. 
The analysts contribute their individual domain knowledge and problem solving abilities across the six steps of BARD and are tasked with producing the best possible solution to the problem the group has been given.

Like the moderator in traditional Delphi processes, the facilitator provides instructions to the group for each step, signals when work is to begin and when responses are to be submitted, reminds analysts of any internal milestones or overall deadlines, presents collated results and other summaries back to the group, and highlights points of interest in those summaries. The collated results are constructed and shared via the dedicated Group workspace at each step, that only the facilitator can edit, but that all analysts can view once they have shared their own work.

The facilitator should incorporate contributions from the analysts to create the group solution, rather than making unnecessary novel contributions themselves. However, the facilitator will make the final decision on what is included and also have editorial control over how it is expressed in the group's final report.

BARD provides two main communication channels:

\ourparagraph{1. Discussion forums.}
There is a separate Discussion Forum for each BARD step. This is where group members can communicate with each other and discuss the challenges for a particular step, gain a better understanding, provide feedback on others' work, and discuss each other's ideas to reach consensus. A new discussion on a particular topic can be started by any analyst or the facilitator. Each topic's discussion is displayed as a single thread, with participants encouraged in the BARD training to use the \texttt{@analyst\_pseudonym} convention to indicate when their comment is a reply to another analyst's comment. 

The provision of a separate Discussion Forum for each step is intended to support the Delphi principle that participants should attempt their own solution before viewing other analysts' contributions; for example, an analyst may be able to read and contribute to a discussion about the BN Variables (Step 2), but if they have not yet provided their attempt at the BN structure (Step 3), then they can't see the Step 3 Discussion Forum. Of course, this relies on the analysts following the protocol and not discussing topics in one forum that are related to a different step. 

\ourparagraph{2. Messaging.} 
BARD's chat message channels provide private, two-way messaging between the facilitator and an analyst, and BARD also allows the facilitator to send a single message to multiple analysts. Messages are not associated with steps. Analysts do not see who else the facilitator may have sent the same message to, and they do not see any messages between the facilitator and other analysts. The message sender may optionally elect to generate an email notification for the recipient(s), which is a useful nudge for someone to login again to BARD when it is being used by the group asynchronously.  However, BARD does not allow analyst-analyst direct messaging, to reduce private ``side'' conversations\footnote{It is difficult to remove all chances of side conversations while enabling public discussion, because analysts could, for example, post their private email addresses onto the forum and set up a conversation outside of BARD.}  and to encourage a collaborative process where all group members have access to the same information and discussions. 

BARD training advocates that the majority of facilitator communication to the group members should be via the Discussion forum. However, the messaging channel is more suitable for contacting individual analysts who have not been contributing either individually or to discussion; for answering private questions from an analyst about using the BARD application (especially where to find help); and for communicating privately to an analyst regarding inappropriate social behaviour (such as showing a lack of respect for others in the discussion forums). The facilitator is supported in these aspects of the role with an administration panel that shows a summary of each group member's last BARD access, and the stage they have reached. 

BARD advocates and supports participants using pseudonyms to maintain their anonymity, a key feature of any Delphi process. However, this aspect is managed by the BARD administrator (see Section~\ref{BARD_tool}) who may choose to have users identified by real names rather than pseudonyms. 

BARD also provides functionality enabling users to automatically incorporate elements of each other's work into their own solution (if they are analysts) or into the Group solution (if they are the facilitator). This is done in slightly different ways in each step, due to the distinct types of content being incorporated. Copying work can greatly reduce the burden on individual analysts, and make it easier to produce compatible contributions. For example, if one analyst likes some of the variables another analyst has defined in step 2, then they can easily adopt them and proceed to demonstrate a slightly different structure in step 3.

The problem solution arising from the BARD process is either (i) the amalgamated consensus Group version produced by the facilitator, or (ii) the highest rated individual BN or Report; BARD supports both options.  

\subsection{The BARD platform}
\label{BARD_tool}

\begin{figure}[ht]
    \centering
    \includegraphics[width=0.45\textwidth]{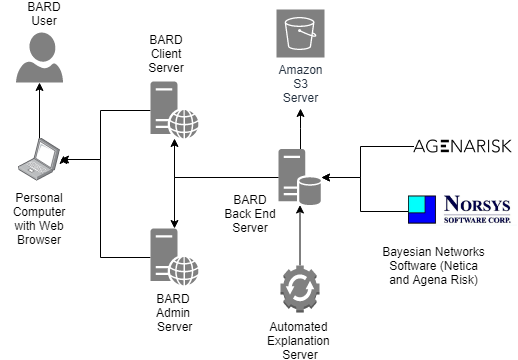}
    \caption{BARD application architecture.}
    \label{fig:bard_arch}
\end{figure}

BARD is a client-server application comprised of a group of cloud-based servers that provide services and resources to connected clients (Figure~\ref{fig:bard_arch}). The main BARD server provides the BARD login, collaboration, problem solving and report generation services, and connects:
(1) a Database server, which provides SQL database services;
(2) a Bayesian Network server, which provides the back-end reasoning via commercial BN software;\footnote{BARD currently runs with both Netica and AgenaRisk BN servers; we anticipate extending to other widely used BN software.}
(3) an Automated Explanation server running the AET (see Step 5 above), which also utilises the BN server; and
(4) a Storage server,\footnote{Currently Amazon Simple Storage Service (S3)}, which stores items such as images uploaded into discussion forums or the report.

The BARD platform has an administrator console, which allows the BARD
administrator to configure the application process flow, enable/disable
certain features (\eg the version of Delphi), and schedule tasks.
The admin console also allows the BARD administrator to manage users, problems and groups, including to: set up problems, optionally with start and end times; create user accounts; create groups to work on a specified problem; allocate users to groups together with (optional) pseudonyms; allocate roles to group members; and download the report from completed problems.

The BARD platform supports an additional role, an Observer, who is assigned to a group and can only observe all stages of the BARD process (in ``read-only'' mode), without being able to contribute to it (apart from messaging the facilitator). They are able to see all public contributions from their group members and facilitator and all steps at any time. This was originally introduced for the CREATE program, to support 'reserve' participants ready to step in as replacements when other participants dropped out of a group during the problem-solving process; but the Observer role has also proved useful for researchers during experiments, and for  analysts and facilitators to keep Observer access for all aspects of a given problem once it ``closes''.

While BARD has been developed as a collaborative BN tool to improve analytical reasoning, stripping out the collaboration features of BARD still leaves a sophisticated analytical tool for solving problems by an individual: we call this version SoloBARD. It allows an analyst to  move through the six steps of BARD without consultation or guidance from anyone else. 

\subsection{BARD Training}
\label{sec:BARD_training}

The BARD Platform comes with approximately 4 hours of training, covering all key elements of the BARD approach:
(1) causal Bayesian network technology;
(2) the BARD workflow including the six steps of BARD and the group interactions;
(3) the BARD software tool, from the perspective of both analysts and the facilitator; and
(4) writing structured analytical reports, using the BARD templates.

This training consists of individual interactive eCourses, produced using the StoryLine 360 tool and hosted on a commercial cloud-based Learning Management System (LMS), called Moodle. The eCourses are all relatively short, from 2--15 minutes, which suits self-paced learning. The LMS allows the BARD training material to be re-packaged into different courses for different purposes, and the training can partitioned. For example, one of the BARD experiments (see below) presented the training courses to its participants in the following partitions: ``required'' (approx 1 hr 30 min), ``recommended'' (approx  1 hr 15 min) and ``optional'' (approx  1 hr 15 min). 

In addition, the LMS is integrated with the BARD software, so that users can visit the LMS at any time from the BARD landing page, and training activity and completion data can be exchanged with BARD, which can then be used for BARD group creation and role allocation.

These LMS-hosted eCourses are further augmented by online help material, which provides assistance and guidance for users within the BARD tool in real time as they work through the BARD process.
These embedded help components include: 
{\em training problems} that allow users to work through elements in the BARD approach as an individual analyst, with pre-populated  ``ideal'' solutions available as the group solution;
an optional {\em product tour} associated with each BARD page, offered on first use and with the ability to revisit;
a {\em general Help facility} that includes PDF and audio-visual versions of the eCourse material; 
{\em context-specific help} as tooltips, page-based tips, and pop-up help tips; and
{\em ``What do I do next?''} guidance.

%% file: evaluation.tex
\vspace*{-2mm}
Here we summarise two experimental studies to test the effectiveness of the BARD approach to problem solving, and their findings; each is reported in detail elsewhere.

\ourparagraph{SoloBARD Experiment \citep{CruzEtAl2019Inprep}.} 
This experiment addressed whether the
BARD system improves individual reasoning on 
three apparently simple probabilistic reasoning problems that each incorporated a tempting qualitative fallacy and also assessed the quantitative accuracy of the answers.
These fallacies are discussed in detail in \cite{PilditchEtAl2018, LiefgreenEtAl2018, PilditchEtAl2019a}.
Individuals in the experimental group (N=29) used the
SoloBARD system, which provides the six Steps of BARD without any of
the social processes, for constructing BNs to use in
reasoning about and solving the problems. The
normative solutions could be achieved via BNs with just seven binary variables. 
The control group
(N=30) received generic training based on the CREATE `Guide to Good
Reasoning' slides and produced their report using MS Office tools. 
External raters
were recruited to blindly assess the final reports from both groups against 
problem-specific marking rubrics. Reasoning was assessed on two measures:
(1)~total rubric score on both qualitative and quantitative questions; (2)~score on only the quantitative questions 
(a subset of the total). On both measures, the group using
SoloBARD performed substantially above controls (see~\cite{CruzEtAl2019Inprep}
for details). These results demonstrate that BARD, even when used
privately by individual analysts, assists them in producing better
reasoned reports for suitable problems. 

\ourparagraph{Groups using BARD End-to-End \citep{KorbEtAl2019Inprep}.} 
This experiment addressed whether, given similar probabilistic reasoning problems to the previous experiment, 
groups using BARD submit better reports than individuals using the best available pen-and-paper tools for probabilistic reasoning.
The experimental condition consisted of groups of up to eight analysts and a facilitator, using the BARD workflow. 
The control group consisted of individuals using Google's online G~Suite tools, frequency formats and chain event graphs \cite[see][]{GigerenzerHoffrage1995}.
Using individuals as the control is an acknowledged limitation of the study as it introduces an additional variable that cannot be distinguished from from the tools used, i.e., group size. However, this was done to match the similar experiment being performed simultaneously by IARPA, and we were unable to add a third condition of groups using G-Suite due to cost constraints and the difficulty of recruiting and retaining enough participants.
External raters were again recruited to assess the final reports from both groups against the problem-specific marking rubrics.
Participants analysed three problems: Problem A in week 1, Problem B in weeks 2 and 3, and Problem C in weeks 4 and 5. The latter two were divided into two stages, partly due to their complexity, but also allowing us to investigate the value of BARD in coping with dynamic problems, 
where evidence and information are updated and require an analysis to be revised. 
198 participants started in the experimental group, with 145 participating in all 5 weeks, while the control group started with 44 and finished with 23 participants. 
The experimental condition outperformed the control
by a significant margin on all problems
(see~\cite{KorbEtAl2019Inprep} for details). This provides evidence that
BARD groups can also beat individuals in producing better
reasoned reports for suitable probabilistic problems, even when the individuals are using the best available pen-and-paper tools. 

In addition, participants from the experimental condition were surveyed at
the end of the experiment to capture feedback on BARD usability, using
the System Usability Scale~\citep{BrookeEtAl1996-sus} to give subjective
usability ratings, and an open-ended questionnaire. Both the ratings
and open-ended comments showed overall positive user satisfaction with
the BARD software, although we note that the results were undoubtedly
skewed in the positive direction because participants who dropped out
of the experiment didn't complete the survey.

%% file: conclusion.tex
We have presented a novel structured technique for collaborative
reasoning and problem-solving that combines a logical procedure for building causal BNs
with a Delphi-style social process. 
BARD is the
first Bayesian network software tool to (1)~break down BN construction
and reasoning into specific steps that guide relatively novice users
through the process, together with minimal upfront training and embedded help, (2)~support groups to collaboratively build a
consensus BN, partly by implementing the entire process in an online platform, and (3)~use the BN to produce a consensus written analytic report,
assisted by a reasoning template and automatically generated key points.
Initial experimental results, summarised in Section~\ref{sec:eval},
are promising for both the usability and effectiveness of the BARD tool for
assisting problem solving and reasoning, by both individuals and groups. The written analytical reports using BARD (29 using BARD individually
in one experiment, 145 using BARD in groups in another) were
significantly better, assessed against problem-specific marking rubrics,
than the controls.

While the version of the BARD tool presented here supports elicitation
of all key elements of a BN, it lacks additional features that are
available in other BN software packages, such as modelling with continuous
variables, learning either the structure or parameters (CPTs) from data,
allowing the CPTs to be specified by equations, supporting decision-making more explicitly
with decision and utility nodes, and sensitivity analysis. We
plan to enhance BARD with these features incrementally, utilising
the functionality of the existing BN software used in BARD's back-end,
subject to resource availability. We are also implementing
functionality to import and export BNs in the formats used by other BN
packages.\footnote{Netica and AgenaRisk, at time of writing.}
Beyond industry-standard BN features, we plan to provide more support for BN idioms, such as those already designed for legal arguments.
There is also a body of work on the
statistical amalgamation of BNs (\eg~\cite{FloresEtAl2011}),
and group elicitation of parameters (\eg~\cite{HaneaEtAl2018-IDEA});
we plan to incorporate some of these methods into BARD for use by the
Facilitator, to further support collaboration.

The BARD platform provides a rich tool for research on how individuals
and groups build and reason with BNs.
BARD's configurable constraint-based structure
layout will allow us to investigate whether particular enforced
BN structure layouts improve understanding of a model and its reasoning,
and aid comparisons between alternative BNs.
We also intend to improve and test the efficacy of the various elements produced by our cutting-edge Automatic Explanation Tool, and improve their presentation by making individual elements available on demand and combining verbal with visual aids.

As we discussed, BARD trades off the rigour of traditional Delphi for the flexibility and user-friendliness of a `real-time' version. Piloting suggested that for our participants and tasks the trade-off is worthwhile.
We also have some experimental evidence that even a minimal Delphi-style interaction improves the network structures produced by BARD groups \citep{BolgerEtAl2019Inprep}.
Nevertheless, we do not yet have any direct experimental comparison for BARD groups between traditional Delphi, real-time Delphi, and free interaction. So, further research comparing social protocols is needed to optimise overall system performance. This investigation will be facilitated by the configurability of BARD user access, with three different versions of Delphi processes already available.
Further possible research on
collaboration includes investigating the best group size
and the factors that may influence it (\eg~\cite{BeltonEtAl2019Inprep}); how a group
may be split across different tasks; and how outputs from multiple groups working in parallel might be considered within a meta-level BARD group.